%% file: mfe.tex
\documentclass{article}
\usepackage{ijcai18}
\usepackage{times}
\usepackage{xcolor}
\usepackage{soul}
\usepackage[utf8]{inputenc}
\usepackage{array}
\usepackage{float}
\usepackage{subfloat}
\usepackage{subcaption}
\usepackage{amsmath}
\usepackage{amsfonts}
\usepackage{amssymb}
\usepackage{color}
\usepackage{bm}
\usepackage[normalem]{ulem}

\usepackage{graphicx}
\graphicspath{{./graphics/}}
\usepackage{mathdesign}
\usepackage{multirow}
\usepackage[T1]{fontenc}    
\usepackage{booktabs}       
\usepackage{breakcites}

\usepackage{caption}
\DeclareCaptionFont{9pt}{\fontsize{9pt}{10pt}\selectfont}
\makeatletter
\let\saved@hyper@linkurl\hyper@linkurl
\let\saved@hyper@link@\hyper@link@
\AtBeginDocument{%
  \NoHyper
  \let\hyper@linkurl\saved@hyper@linkurl 
  \let\hyper@link@\saved@hyper@link@ 
}
\makeatother

\makeatletter

\newcommand{\Rmnum}[1]{\expandafter\@slowromancap\romannumeral #1@}
\makeatother

\title{Medical Concept Embedding with Time-Aware Attention}

\author{
Xiangrui Cai$^1$\thanks{The majority of this work was completed while the 1st author was visiting Naitional University of Singapore.}, 
Jinyang Gao$^2$, 
Kee Yuan Ngiam$^3$, 
Beng Chin Ooi$^2$,
Ying Zhang$^1$,
Xiaojie Yuan$^1$
\\ 
$^1$ Nankai University, China\\
$^2$ National University of Singapore, Singapore\\
$^3$ National University Health System, Singapore\\
\{caixiangrui, zhangying, yuanxiaojie\}@dbis.nankai.edu.cn,\\
\{jinyang.gao, ooibc\}@comp.nus.edu.sg,~
kee\_yuan\_ngiam@nuhs.edu.sg\\
}

\begin{document}

\maketitle

\begin{abstract}

	Embeddings of medical concepts such as medication, procedure and diagnosis codes in Electronic Medical Records (EMRs) are central to healthcare analytics. Previous work on medical concept embedding takes medical concepts and EMRs as words and documents respectively. Nevertheless, such models miss out the temporal nature of EMR data. On the one hand, two consecutive medical concepts do not indicate they are temporally close, but the correlations between them can be revealed by the time gap. On the other hand, the temporal scopes of medical concepts often vary greatly (e.g., \textit{common cold} and \textit{diabetes}). In this paper, we propose to incorporate the temporal information to embed medical codes. Based on the Continuous Bag-of-Words model, we employ the attention mechanism to learn a ``soft'' time-aware context window for each medical concept. Experiments on public and proprietary datasets through clustering and nearest neighbour search tasks demonstrate the effectiveness of our model, showing that it outperforms five state-of-the-art baselines.

\end{abstract}

\input{introduction.tex}
\input{related_work.tex}
\input{model.tex}

\input{experiments.tex}

\section{Conclusions} 

We introduced a model that learned the embedding and a ``soft'' temporal scope for each
medical concept simultaneously. Based on the CBOW model, our model takes advantage of attention
mechanisms to learn such temporal scopes.
The experimental results on two datasets and two tasks demonstrate the
effectiveness of our models compared to state-of-the-art models. Our next plan
is to utilize both medical concept embeddings and the ``soft'' context scopes
for healthcare tasks such as missing value imputation.


\section*{Acknowledgments}
This research is supported by National Natural Science Foundation of China (No.
61772289) and National Research Foundation, Prime Ministers Office, Singapore
under CRP Award (No. NRF-CRP8-2011-08). We thank the four anonymous reviewers
for their valuable comments on our manuscript. We also thank Sihan Xu for her
suggestions on the organization of this paper.

\bibliographystyle{named}
\bibliography{refs}
\end{document}

%% file: introduction.tex
\section{Introduction} \label{intro}

The rapid growth in use of Electronic Medical Records (EMRs) offers promises
for healthcare analytics~\cite{hillestad2005can}, such as chronic disease
management and personalized medicine. EMRs contain a wealth of healthcare
information, including medication, procedure, diagnosis codes and lab test
results. For clinical and educational purposes, these medical concepts have
been standardized by well-organized ontologies such as the International
Classification of Diseases (ICD) \footnote{\href{http://www.who.int/classifications/icd/en/}{http://www.who.int/classification/icd/en}} and the National Drug Code (NDC)
\footnote{\href{https://www.fda.gov/Drugs/InformationOnDrugs/ucm142438.htm}{http://www.fda.gov}}.

Representations for such medical codes is at the heart of healthcare
analytics. One way is to use one-hot vectors to represent medical codes.
However, the one-hot vectors are naturally high-dimensional, sparse and can hardly
represent the semantic relationships of medical concepts. There has also been some existing work that designs hand-crafted features for healthcare predictive models~\cite{sun2012supervised,ghassemi2014unfolding}. Due to the requirement for expert knowledge, the scalability of these methods is limited.

The use of distributed representations is motivated by the successful
applications in natural language processing~\cite{bengio2003neural,collobert2008unified,mikolov2013distributed,pennington2014glove}. Recently, a variety of
embedding approaches have emerged as effective methods to learn representations
of medical concepts~\cite{minarro2014exploring,de2014medical,tran2015learning,choi2016learning,choi2016multi,choi2017gram}. They all treat medical concepts
and EMRs as words and documents respectively. The basic idea behind these
models is that similar words (medical concepts) may share similar
contexts~\cite{harris1954distributional}. Hence, the key challenge of medical
concept embedding is how to represent the contexts of medical concepts
effectively without loss of information.

\begin{figure}[h]
	\begin{center}
		\includegraphics[scale=0.55]{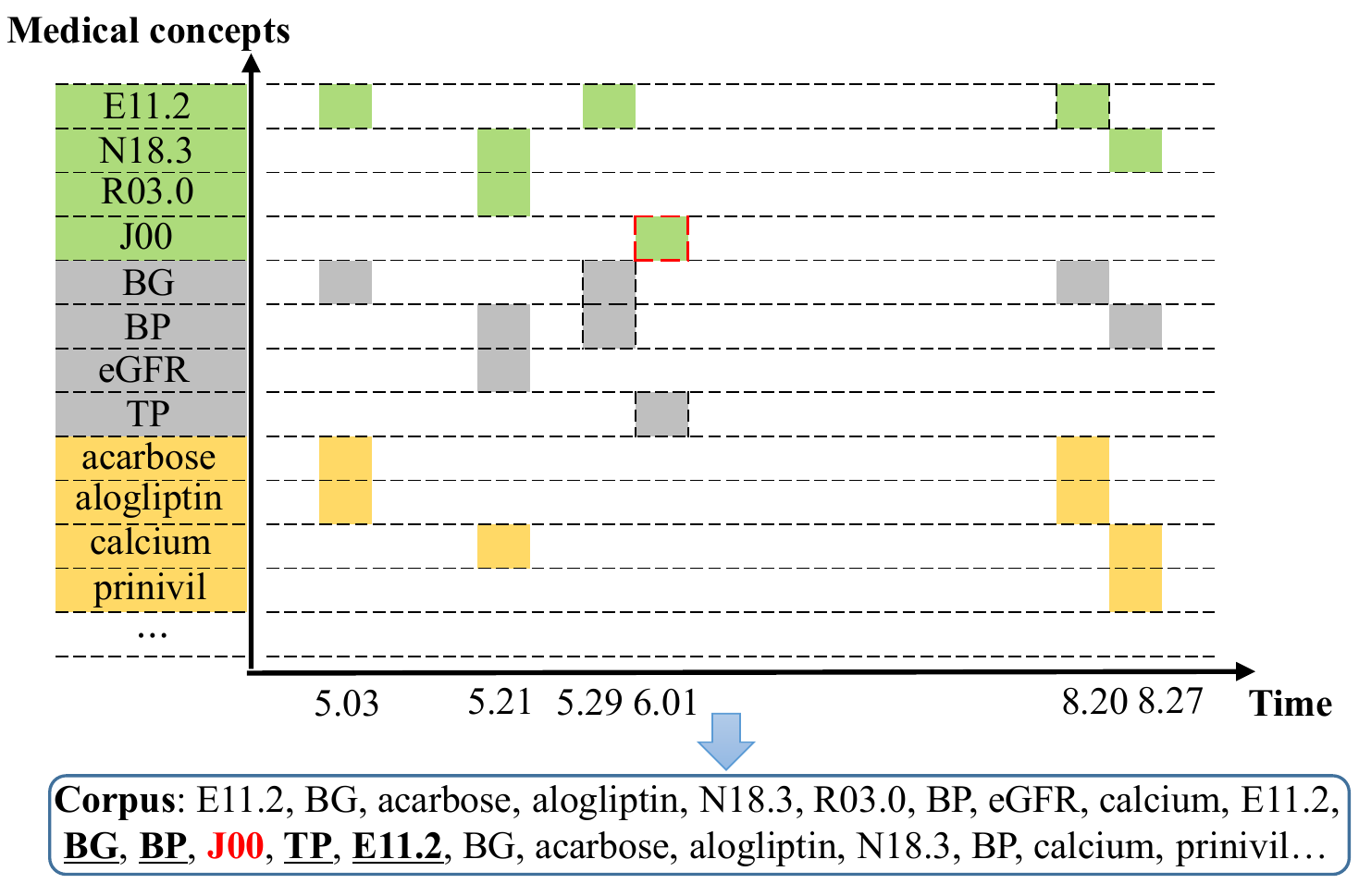}
	\end{center}
  \vskip -1em
	\caption{An EMR segment and its corpus. The squares indicate that the medical codes are recorded at the time stamps.}
	\label{fig:emr}
\end{figure}

Nevertheless, current methods only consider code co-occurrences within a
fixed-size window as indications of contexts. The temporal nature of EMRs has
been ignored. Figure~\ref{fig:emr} illustrates a segment of a patient's EMR
data. Given the window size to be 2, conventional methods equally take
\textit{Blood Glucose} (BG, May 29), \textit{Blood Pressure} (BP, May 29),
\textit{Temperature} (TP, Jun 1) and \textit{Diabetes mellitus} (E11.2, Aug 20)
as the contexts of \textit{Common cold} (J00, May 29). However, the time gap
between \textit{J00} (Jun 1) and \textit{E11.2} (Aug 20) is more than two
months, indicating that these two concepts may not be regarded as contexts of
each other. Due to irregular visits of patients, this is very common in EMRs.

One straightforward solution is to define a fixed-size temporal scope to derive
code co-occurrences. However, we observe that the scopes of medical codes often
vary greatly. For example, \textit{common cold} often lasts for weeks, while
\textit{diabetes mellitus} typically lasts for years~\cite{chiang2014type}.
Therefore, a fixed-size temporal window can hardly satisfy all medical concepts.
To tackle the aforementioned challenges, this paper proposes to learn the
embeddings and temporal scopes of medical concepts simultaneously. Instead of
directly learning a ``hard'' temporal scope for each medical concept, which is
impractical due to the exponential complexity, we learn medical concept
embeddings and relations between medical concepts and consecutive time periods
simultaneously. Based on the Continuous Bag-of-Words model
(CBOW)~\cite{mikolov2013efficient}, we build a time-aware attention model to
learn such ``soft'' temporal scopes. Each attention weight describes the
contribution of medical concepts within a certain time period to predict the
central medical concept.

Generally, the contributions of this paper can be summarized as follows.

\begin{itemize}
  \item We observe the temporal information in EMRs that benefits capturing co-occurred medical concepts. Furthermore, we also find that temporal scopes of medical concepts often vary greatly.
  \item We propose a time-aware attention model based on CBOW. The model learns
  representations and ``soft'' temporal scopes of medical concepts
  simultaneously.
  \item We evaluate the effectiveness of the proposed model on two datasets with clustering and nearest neighbor search tasks. The results show that our model outperforms five state-of-the-art models.
\end{itemize}

%% file: related_work.tex
\section{Related Work}
The earliest idea of word embeddings dated back to
1988~\cite{earlistembedding}. \cite{bengio2003neural} proposed a neural network
model to learn word embeddings, followed by a surge of models
~\cite{collobert2008unified,mikolov2013efficient,mikolov2013distributed,
pennington2014glove,bojanowski2016enriching}. All of them shared the same
principle that words appeared in similar contexts should be
similar~\cite{harris1954distributional}. Recently, some work has explored the
influence of context scopes for neural word embeddings. \cite{melamud2016role}
investigated the influences of context types for the Skip-gram model.
\cite{liu2017context} modeled the probability of the target word conditioned on
a subset of the contexts based on the exponential-family embedding model (EF-
EMB)~\cite{rudolph2016exponential}. Closest to our approach, \cite{ling2015not}
introduced an attention model to consider contextual words differently based on
the CBOW model.

Following the same idea of word2vec models
\cite{mikolov2013efficient,mikolov2013distributed}, recent work has applied the
models directly to the healthcare domain. \cite{minarro2014exploring} applied
the Skip-gram model to medical texts collected from PubMed, Merck Mannuals,
etc. \cite{de2014medical} took the same model to learn embeddings of UMLS
concepts from medical notes and journal abstracts. \cite{choi2016learning}
employed the Skip-gram model to learn the representations of medical concepts
from medical journals, medical claims and clinical narratives separately. These
work was very close to word embeddings, which was trained on documents.
\cite{tran2015learning} embedded medical concepts with an non-negative
restricted boltzmann machine in an EMR data set. \cite{choi2016multi} proposed
\textit{med2vec} to learn representations of both features and visits in EMR
data based on the Skip-gram model. \cite{choi2017gram} learned task-oriented
embeddings of medical concepts and proposed to incorporate well-organized
ontologies. All these work treated EMRs in the same way as documents,
ignoring the time information. None are aware of the various context
scopes of medical concepts, which in return is a main challenge for medical concept embedding.

Medical concept embeddings have been shown to be beneficial to 
 downstream applications, such as predicting diagnoses and patient status 
 \cite{pham2016deepcare,che2017exploiting}, discovering 
 clinical motifs \cite{nguyen2017mathtt} and identifying similar patients 
 \cite{zhu2016Measuring,suo2017personalized}. 



%% file: model.tex
\section{Methodology}
In this section, we first briefly review the CBOW model, and then describe our proposed method in detail.

\subsection{CBOW}
Given a
training corpus, represented as a word sequence $W =\left \{ w_1, w_2, \dots, w_N \right \}$, The CBOW model~\cite{mikolov2013efficient} learns word representations by using the context words within a sliding window to predict the target word. This is achieved by maximizing the average
log probability of the occurrences of target words given context words:
\begin{equation}
	\label{eq:cbow}
	\frac{1}{N}\sum_{n=L}^{N-L} \log p(w_n|H_n) \,,
\end{equation}
where $L$ refers to the size of the context window and
$H_n=\{w_{n-L},\dots,w_{n-1},w_{n+1},\dots,w_{n+L}\}$. The CBOW model learns two
vectors for each word. One input vector $\bm v_{w_n}$ represents $w_n$ as a
context word, and one output vector $\bm v_{w_n}^\prime$ represents $w_n$ as a
target word. The hidden representation of context words is obtained by averaging
their input vectors, i.e., $\bm h_n = \frac{1}{2L}\sum_{w_j \in H_n}\bm v_{w_j}$. The conditional
probability in Equation \eqref{eq:cbow} is modeled by a softmax function:
\begin{equation}
	\label{eq:softmax}
	p(w_n | H_n) = \frac{\exp \{{\bm v_{w_n}^\prime}^\top \bm h_n\}}{\sum_{w=1}^{|V|} \exp \{{\bm v_w^\prime}^\top \bm h_n\}} \,,
\end{equation}
where the word vocabulary is $V$. To reduce the computational complexity of optimization, the CBOW model uses negative sampling to
approximate the softmax formulation, which instead maximizes:
\begin{equation}
	\label{eq:ns}
	\log \sigma({\bm v_{w_n}^\prime}^\top \bm h_n) + \sum_{i=1}^r \mathbb{E}_{w_x \sim P(w)} [\log \sigma(-{\bm v_{w_x}^\prime}^\top \bm h_n)] \,,
\end{equation}
where $\sigma$ is the sigmoid function, and $r$ is the number of negative
samples. Each negative sampled context $w_x$ follows the unigram distribution
raised to 3/4th power, i.e., $P(w) = U(w)^{\frac{3}{4}}/A$ ($A$ is a constant)
\cite{mikolov2013distributed}. 
The computational complexity is proportional to $r$, which is
usually small, e.g., $r=5$.

\subsection{Medical Concept Embedding}

Medical concepts in EMR data are associated with time stamps. As introduced
in Section~\ref{intro}, the temporal information in EMR data can benefit medical concept embedding by improving the hidden representations of contexts. Considering the various scopes of medical concepts, we propose a time-aware attention model to learn better embeddings of medical concepts based on the CBOW model.

\subsubsection{Model Architecture}

Given a longitudinal EMR data with a finite set of medical concepts $C$,
medical concept embedding aims at learning distributed representations for
medical concepts. An EMR sequence consists of several visits of a patient to the
hospital, where the visits are ordered temporally and each of them is composed
of multiple medical concepts. That is, an EMR sequence can be represented by a
sequence of medical concept subsets, where each subset contains medical concepts
with the same time stamps. Formally, by defining a time unit, such as a day, a
week, a month, etc., an EMR sequence is denoted by $E = (E_1, E_2, \dots, E_T)$,
where $E_t$ is a subset of medical concepts within the $t$th time unit ($E_t
\subset C$). We represent $E_t$ by $E_t = \{c_{t,1}, c_{t,2},\dots, c_{t,
K_t}\}, c_{t,i} \in C$, where $c_{t,i}$ denotes the $i$th medical concept of
$E_t$ and $K_t$ the total number of medical concepts in $E_t$. 

\begin{figure}
  \centering
\includegraphics[scale=0.6]{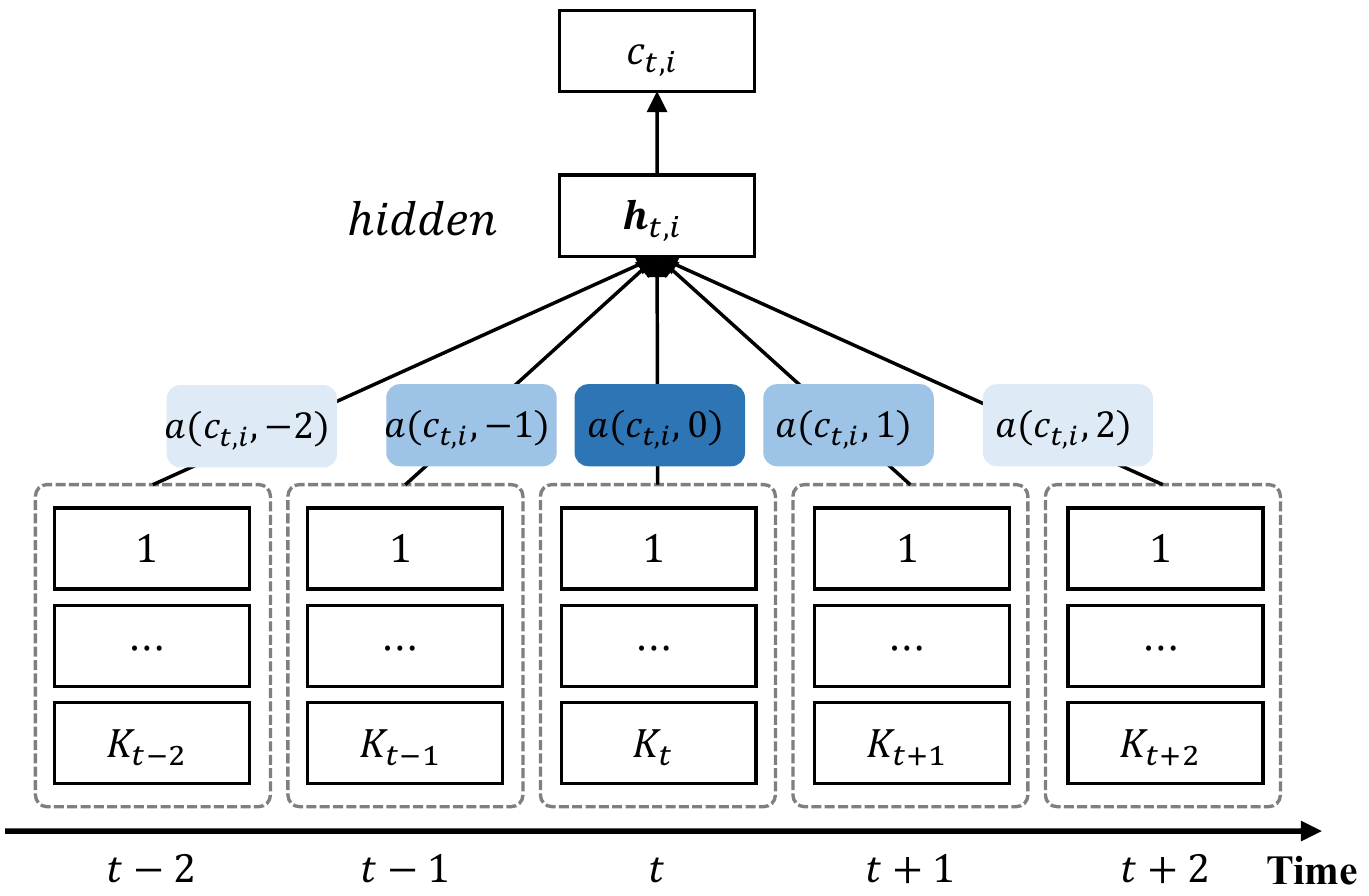}
\caption{Time-aware attention model for medical concept embedding based on CBOW;
the darker cells indicate heavier influence of $c_{t,i}$ on the corresponding
time unit.}
\label{fig:model}
\end{figure}

Our medical concept embedding model, abbreviated as MCE, is based on the CBOW
model. As shown in Figure \ref{fig:model}, the MCE model predicts a target
medical concept using its surrounding contexts within a sliding window.  Similar
to CBOW, a medical concept $c_{t,i}$ in MCE is represented by two vectors, one
input vector to represent it as a context word, and one output vector to
represent it as a target word, which are denoted by $\bm v_{c_{t,i}}$ and $\bm
v_{c_{t,i}}^\prime$ respectively. The MCE model takes advantage of negative
sampling to approximate the following objective of each target-contexts pair:
\begin{equation}
	\log \sigma({\bm v_{c_{t,i}}^\prime}^\top \bm h_{t,i}) + \sum_{x=1}^r \mathbb{E}_{c_x \sim P(c)} [\log \sigma(-{\bm v_{c_x}^\prime}^\top \bm h_{t,i})] \,.
\end{equation}
where $c_{t,i}$ refers to the target medical concept and $\bm h_{t,i}$ the
hidden representation of the contexts of $c_{t,i}$. The negative sample $c_x$ is
obtained in the same way as the CBOW model.

\subsubsection{Time-aware Attention}

Considering the temporal nature of EMR data, we observe that the medical
concepts vary greatly in terms of temporal scopes. One solution is to optimize
the best temporal scope for each medical concept, which is impractical due to
the exponential complexity. We notice that relations between medical concepts
and time periods can be learned from the correlations between medical concepts.
For example, \textit{common cold} is usually related to medical codes that
appears within one week from it, which indicates the \textit{common cold} has
larger influence on one week than the other far away periods. To this end, we
build a time-aware attention model to learn non-uniform attention weights within
a temporal context scope. The temporal context scope is defined as the largest
number of time units between the contexts and the target medical concepts. The
non-uniform attentions on different time periods are regards as ``soft'' context
scopes in this paper.

Specifically, we first split an EMR sequence by time units. Medical concepts
within a time unit $t$ are represented by $E_t$. Given a temporal context scope $S$, the contexts of a medical concept $c_{t,i}$ is denoted by
$H_{t,i}=\{E_{t-S},E_{t-S+1},\dots,E_{t-1},E'_t,E_{t+1},\dots,E_{t+S} \}$, where
$S$ denotes the temporal scope and $E'_t = \left \{ c_{t,j}|c_{t,j}\in E_t
\wedge j\neq i\right \}$. Moreover, we find that there could be large
amounts of medical concepts within a visit. To avoid context explosion, which
leads to highly computational costs, we limit the largest number of context
medical concepts by a threshold $\Gamma$, i.e., $|H_{t,i}| \le \Gamma$. We will
discuss the effects of the temporal window size $S$ and the context threshold
$\Gamma$ in Section~\ref{exp}. The hidden representation is composed by non-uniform
weighted context vectors, which incorporates the time-aware attention model: 
\begin{equation}
\label{eq:context_vector_f}
\bm h_{t,i} = \sum_{c_{q,j} \in H_{t,i}} a(c_{t,i}, \Delta_q) \bm v_{c_{q,j}} \,,
\end{equation}
where $\Delta_q=q-t$ and $a(c_{t,i}, \Delta_q)$ models an attention level given
to contexts within $E_{\Delta_q}$ for predicting the target medical concept
$c_{t,i}$. It is parameterized by the target medical concept and the 
time gap between each context and the target, i.e.,
\begin{equation}
\label{eq:attn_context}
a(c_{t,i}, \Delta_q) = \frac{\exp \{ m_{c_{t,i},\Delta_q} + b_{\Delta_q} \}} {\sum_{c_{q^\prime,j^\prime} \in H_{t,i}} \exp \{m_{c_{t,i}, \Delta_{q^\prime}} + b_{\Delta_{q^\prime}}\}} \,,
\end{equation}
where $\bm m \in \mathbb R^{|C| \times (2S+1) }$ and $\bm b \in \mathbb
R^{2S+1}$. Each $m_{c,\Delta}$ determines the influence of the medical concept
$c$ on relative time period $\Delta$. $\bm b$ is a bias, conditioning on the
relative time period only. Taking \textit{common cold} as an example again, the
model finds most contexts within a week are related to \textit{common cold}.
Then the model learns to pay more attentions on this time period (one week),
which could be used in next iteration to denoise the irrelevant contexts that
are far behind \textit{common cold}. Hence, our model can improve medical
concept embeddings by identifying the related regions and capturing more
accurate related target-context pairs.



\subsubsection{Parameter Learning}

The learning algorithm for our model is stochastic gradient descent. We compute
gradients of the loss with regards to the parameters with back-propagation. The
learning rate is decayed linearly, as the same for training CBOW.

Compared to the CBOW model, the overhead of our model is only the
additional computation of attentions. Each operation of computing an attention
weight is to fetch two parameters from $\bm m$ and $\bm b$. Hence, the
complexity of computing the attentions is proportional to the number of
contexts, which is limited by the context threshold $\Gamma$.

%% file: experiments.tex
\section{Experiments}\label{exp}
In experiment, we evaluate the performance of the proposed model on public and proprietary datasets through clustering and nearest neighbour search tasks. The source code is available at https://github.com/XiangruiCAI/mce.

\subsection{Datasets}
We perform comparative experiments on both public and proprietary datasets listed as follows:

\textbf{NUH2012} is a real-world dataset provided by National University
Hospital of Singapore. This dataset contains EMR data spanning 2012. We extract
diagnosis codes, medications, procedures and laboratory test results from this
dataset to perform experiments. The diagnosis codes follows the 10th revision of
the ICD (ICD-10).
\footnote{\href{http://www.icd10data.com/ICD10CM/Codes}{http://www.icd10data.com/ICD10CM/Codes}}

\textbf{DE-SynPUF} \footnote{\href{https://www.cms.gov/Research-Statistics-Data-and-Systems/Downloadable-Public-Use-Files/SynPUFs/DE_Syn_PUF.html}{\color{black}{https://www.cms.gov}}} 
is a public dataset provided by Centers for Medicare and Medicaid Services (CMS). This dataset contains three years of data (2008-2010), providing inpatient, outpatient and carrier files, along with the beneficiary summary files. Although some variables have been synthesized to minimize the risk of re-identification, the sheer volume of EMR data can still provide useful insights for medical data processing. The diagnosis codes in this dataset follow the ICD-9
standard. \footnote{\href{http://www.icd9data.com/}{http://www.icd9data.com}}

We present the statistics of both two datasets in Table \ref{tbl:stat}. It can be observed that DE-SynPUF is a large-scale dataset with more patients, time stamps and unique medical concepts compared to NUH2012. Datasets with different scales are utilized to demonstrate the validity and compatibility of the proposed model.

\begin{table}
\centering
\begin{tabular}{lcc}
\hline
\toprule
Data set & NUH2012 & DE-SynPUF \\
\midrule
\#patients & 30758 & 1984582 \\
\#time stamps & 801314 & 106477456 \\
\#unique medical concepts & 13382 & 34491 \\
\#time stamps / patient & 26.0522 & 53.6523 \\
\#codes / time stamp & 6.4164 & 6.3908 \\
max \#codes / time stamp & 127 & 78 \\
size after preprocessing & 237MB & 5.4GB \\
\bottomrule
\end{tabular}
\caption{The statistics of two datasets.}
\label{tbl:stat}
\end{table}
\subsection{Ground Truth}

We perform the clustering and nearest neighbour search (NNS) tasks to evaluate
the quality of our medical concept embeddings. The ground truth is derived from
two well-organized ontologies, i.e., the ICD standard and Clinic Classifications
Software
(CCS)\footnote{\href{https://www.hcup-us.ahrq.gov/}{https://www.hcup-us.ahrq.gov}}.
The ICD standard has a hierarchical structure and we use the root-level nodes as
the clustering labels. Medical concepts under the same root are expected to be
clustered into the same group. For nearest neighbour search, we take
medical concepts under the same subroot as near neighbours. We obtain 21
categories and 755,000 near neighbour pairs for NUH2012; we also derive 19
categories and 4,677,706 near neighbour pairs for DE-SynPUF. This set of ground
truth is named by Hier in the following sections.

Considering that medical concepts in different ICD subtrees can be semantic
related in some cases, we take CCS as complementary clinic classifications that
involve expert knowledge. We note that in CCS, both ICD-9 and ICD-10 standards
have the same 285 categories, which are used to assess the quality of clustering
for both datasets. Similar to Hier, medical concepts that appear in the same
sub-category are regarded as near neighbors in CCS. Finally, we obtain 53,812
pairs of nearest neighbour for NUH2012 and 2,152,632 pairs for DE-SynPUF. We
refer to this set of ground truth as CCS.

\subsection{Baselines and Training Details}\label{parameter}

We compare our model against 5 state-of-the-art models, i.e., Skip-gram, CBOW~\cite{mikolov2013efficient,mikolov2013distributed},
Glove~\cite{pennington2014glove}, wang2vec~\cite{ling2015not} and
med2vec~\cite{choi2016multi}. All these models have been trained with their source
code.

For preprocessing the datasets, medical concepts occurred less than 5 times are discarded for all models. During training, the rejection threshold is $10^{-4}$. We use the same negative sampling estimation for CBOW, Skip-gram, wang2vec and our model, and the number of negative samples is 5. The start learning rate is set as 0.05 and 0.025 for Skip-gram and the CBOW-based models respectively. For the proposed model, we set time unit as one week. We note that the maximum number of contexts in our model is set as twice as the context window size $W$ in the baselines, within which the number of contexts is typically $2W$. All models are trained with 30 epochs for NUH2012 and 5 epochs for DE-SynPUF. The dimension of medical concept vectors is 100 for all models.

\subsection{Results}
 
 We perform comparative studies on both public and proprietary datasets through the clustering and nearest neighbour search tasks. To evaluate the performance on clustering, we apply k-means to the learned medical concept embeddings. As mentioned in section~\ref{parameter}, we set the maximum number of contexts in our model to be 60, twice as large as the window size in the baselines. The temporal scope of our model is 20 (weeks). Two sets of ground truth, i.e., Hier and CCS, are both utilized to evaluate the performance of these models. We note that the results of med2vec on the DE-SynPUF dataset can not be obtained after two weeks of training.

 Table \ref{tbl:nmi} reports the normalized mutual information (NMI) for clustering, and Table \ref{tbl:precision} presents precision@1 (P@1) for NNS. The best
results are \textbf{bold-faced} in these tables. From Table \ref{tbl:nmi} and Table \ref{tbl:precision}, we can come to a conclusion that the MCE model outperforms most state-of-the-art models on medical concept embedding. The performance gains of MCE over the other models indicate that, by incorporating the temporal information in EMR data, better medical concept representations can be learned. In addition, among all these models, CBOW and Skip-gram are still competitive to MCE, whereas the performance of med2vec is observed to be the worst on both tasks. For med2vec, we observe that medical concepts with low frequencies are assigned near-zero vectors in this model, which might explain for its poor performance.

\begin{table}
\centering
\begin{tabular}{lcccc}
\hline
\toprule
\multirow{2}{*}{Model} & \multicolumn{2}{c}{NUH2012} & \multicolumn{2}{c}{DE-SynPUF} \\ [2pt]
\cline{2-5}\rule{0pt}{12pt}
& Hier & CCS & Hier & CCS \\
\midrule
CBOW      & 33.73 & 63.20 & 47.17 & 90.33 \\
Skip-gram & 33.22 & 63.96 & 50.12 & 90.36 \\
wang2vec  & 28.35 & 59.35 & 53.71 & 91.66 \\
Glove     & 17.41 & 55.34 & 45.27 & 87.57 \\
med2vec   & 7.51  & 55.52 &   -      & -         \\
MCE       & \textbf{35.09} & \textbf{65.46} & \textbf{54.96} & \textbf{93.18} \\
\bottomrule
\end{tabular}
\caption{Clustering performance (NMI) of the models on two datasets w.r.t.
ground truth Hier and CCS (\%).}
\label{tbl:nmi}
\end{table}

\begin{table}
\centering
\begin{tabular}{lcccc}
\hline
\toprule
\multirow{2}{*}{Model} & \multicolumn{2}{c}{NUH2012} & \multicolumn{2}{c}{DE-SynPUF} \\ [2pt]
\cline{2-5}\rule{0pt}{12pt}
& Hier & CCS & Hier & CCS \\
\midrule
CBOW      & 14.69 & 29.91 & 31.87 & 88.35 \\
Skip-gram & 14.87 & 30.01 & 32.63 & 91.41 \\
wang2vec  & 10.18 & 15.40 & 32.90 & 93.20 \\
Glove     & 9.08  & 15.61 & 31.98 & 86.72 \\
med2vec   & 3.09  & 4.86  & -        & -        \\
MCE       & \textbf{17.74} & \textbf{32.85} & \textbf{33.88} & \textbf{94.84} \\
\bottomrule
\end{tabular}
\caption{NNS performance (P@1) of the models on two datasets w.r.t. ground truth Hier and CCS (\%).}
\label{tbl:precision}
\end{table}

 Moreover, by comparing the results on two datasets and two sets of ground truth, more useful findings can be obtained from Table \ref{tbl:nmi} and Table \ref{tbl:precision}. First, with the same experimental settings, all listed models achieve better performance on CCS than Hier. Especially for Glove, there is a wide gap between the results on Hier and CCS. This trend might be explained by the expert knowledge involved in CCS. Second, we can observe that models trained with DE-SynPUF perform better than those trained with NUH2012. A possible reason could be that although containing synthesized variables, DE-SynPUF has a larger scale than NUH2012, which provides richer information for medical data processing than the small dataset.

\textbf{Performance variance with different window sizes}. To make fair comparisons, we investigate the performance of these models by varying the context window size. For there is no parameter for window size in med2vec, we only compare the proposed model against four baselines. Specifically, we vary the window size from $10$ to $100$ for each model. As mentioned in section~\ref{parameter}, the context threshold for MCE is set twice as large as the window size in the baselines.

\begin{figure*}[ht]
\centering
\includegraphics[scale=0.35]{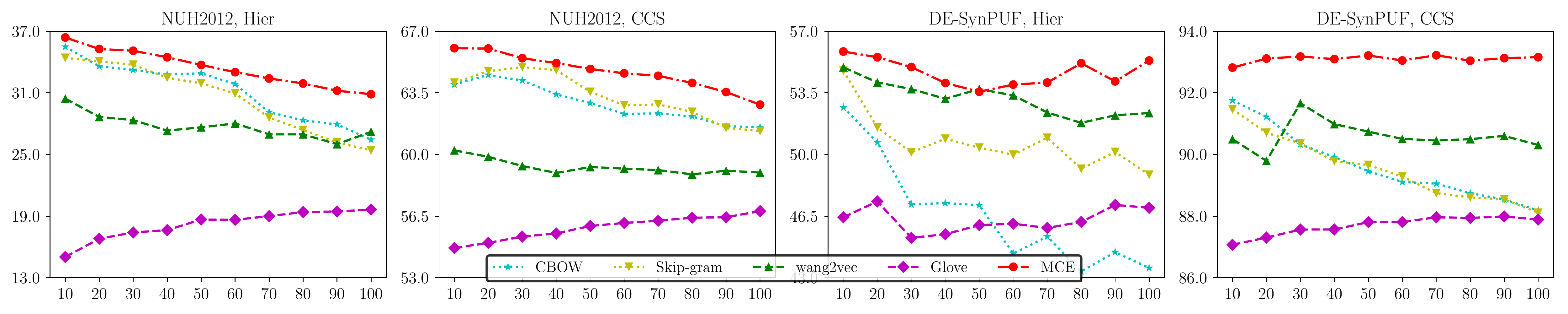}
\caption{Clustering results on two datasets and two ground truth in terms of NMI
(\%). The window size varies from 10 to 100.}
\label{fig:varied_ws_NMI}
\end{figure*}

We summarize the results on the clustering task in Figure \ref{fig:varied_ws_NMI}. Generally, we can observe that as the window size (context threshold) increases, most models decrease their performance due to the induced noise. Nevertheless, the MCE model always outperforms the rest models in terms of NMI. Specifically, as we increase the window size, the performance of CBOW and Skip-gram decreases significantly, while Glove achieves better performances with larger window size. A possible explanation is that Glove takes use of global co-occurrences, which might benefit from the large window size. The results of wang2vec also get worse as the window size increases. Compared to these models, it can be seen that although getting slightly worse on NUH2012, the MCE model achieves nearly stable performances on DE-SynPUF, which shows its scalability when trained with large-scale datasets.

\begin{figure*}
\centering
\includegraphics[scale=0.35]{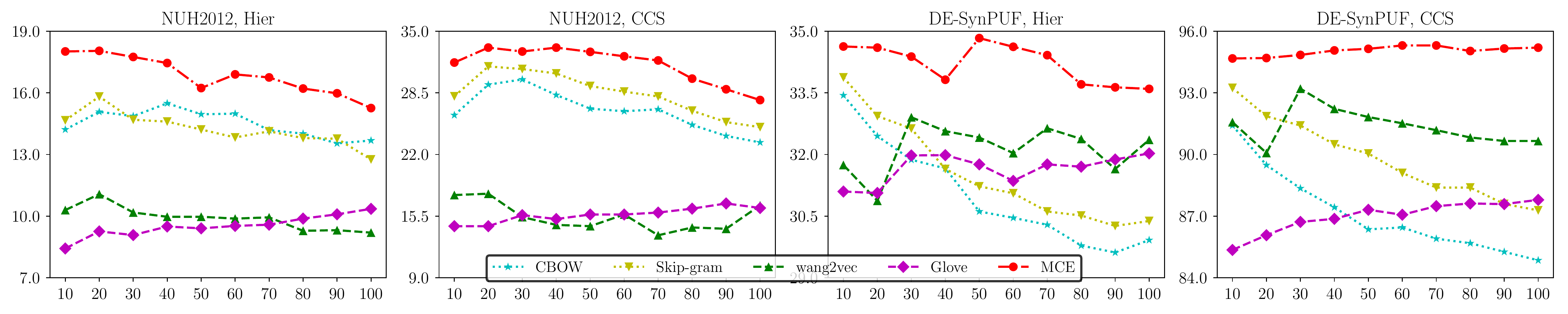}
\caption{NNS results on two datasets and two ground truth in terms of P@1 (\%).
The window size varies from 10 to 100.}
\label{fig:varied_ws_P}
\end{figure*}

Similar scenario can be found in Figure \ref{fig:varied_ws_P}. The proposed
MCE model outperforms the rest competitors, and achieves stable performance on a
large-scale EMR dataset DE-SynPUF. The performance of the CBOW and Skip-gram model decreases significantly on DE-SynPUF as the window size increases, while Glove outperforms these two models when trained with a large window size.


\subsection{Effects of the Temporal Scope}

\begin{figure}
  \centering
\begin{subfigure}[t]{0.5\linewidth}
\centering
  \includegraphics[width=\linewidth]{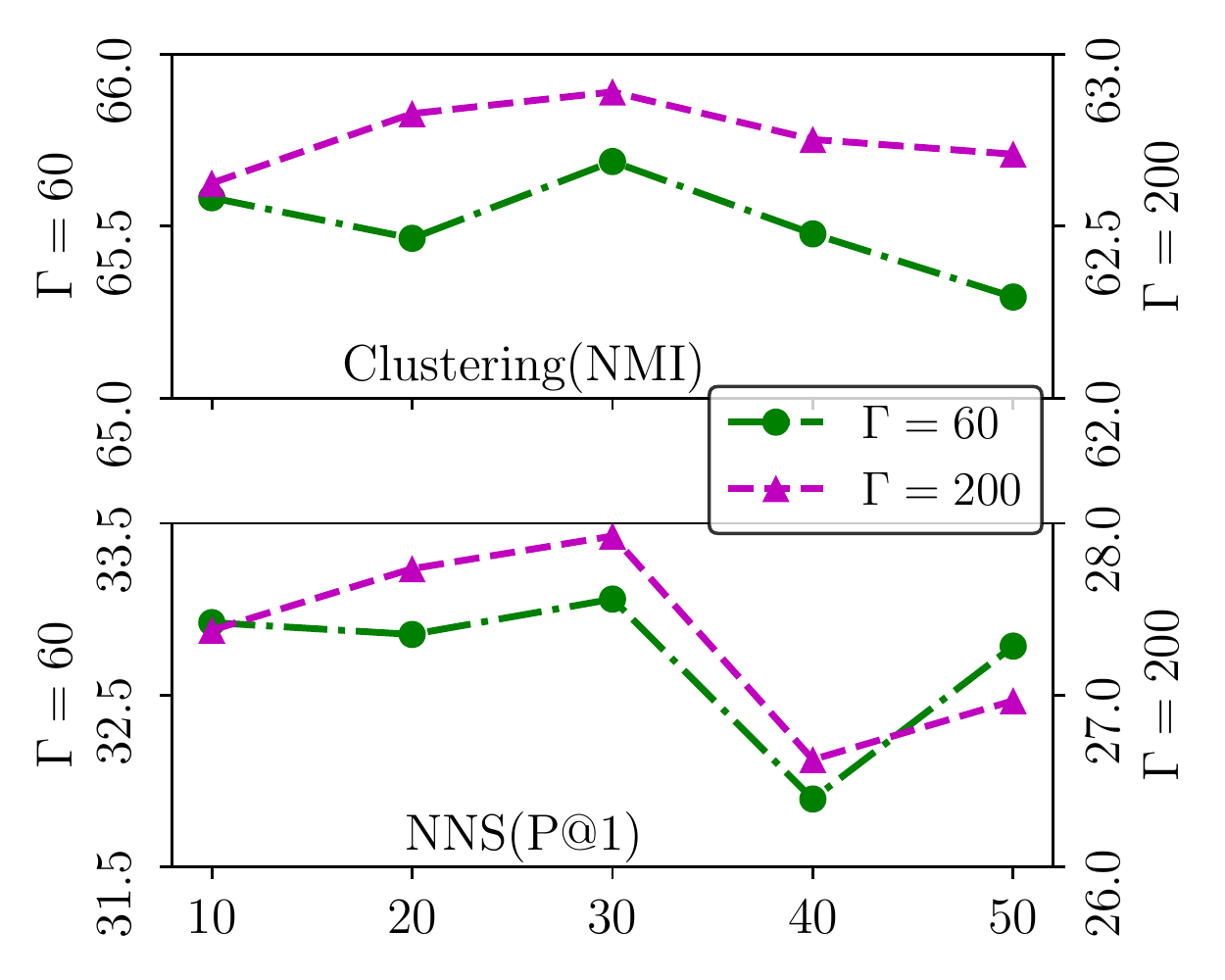}
  \caption{NUH2012}\label{fig:varied_attn_nuh}
\end{subfigure}%
\begin{subfigure}[t]{0.5\linewidth}
\centering
  \includegraphics[width=\linewidth]{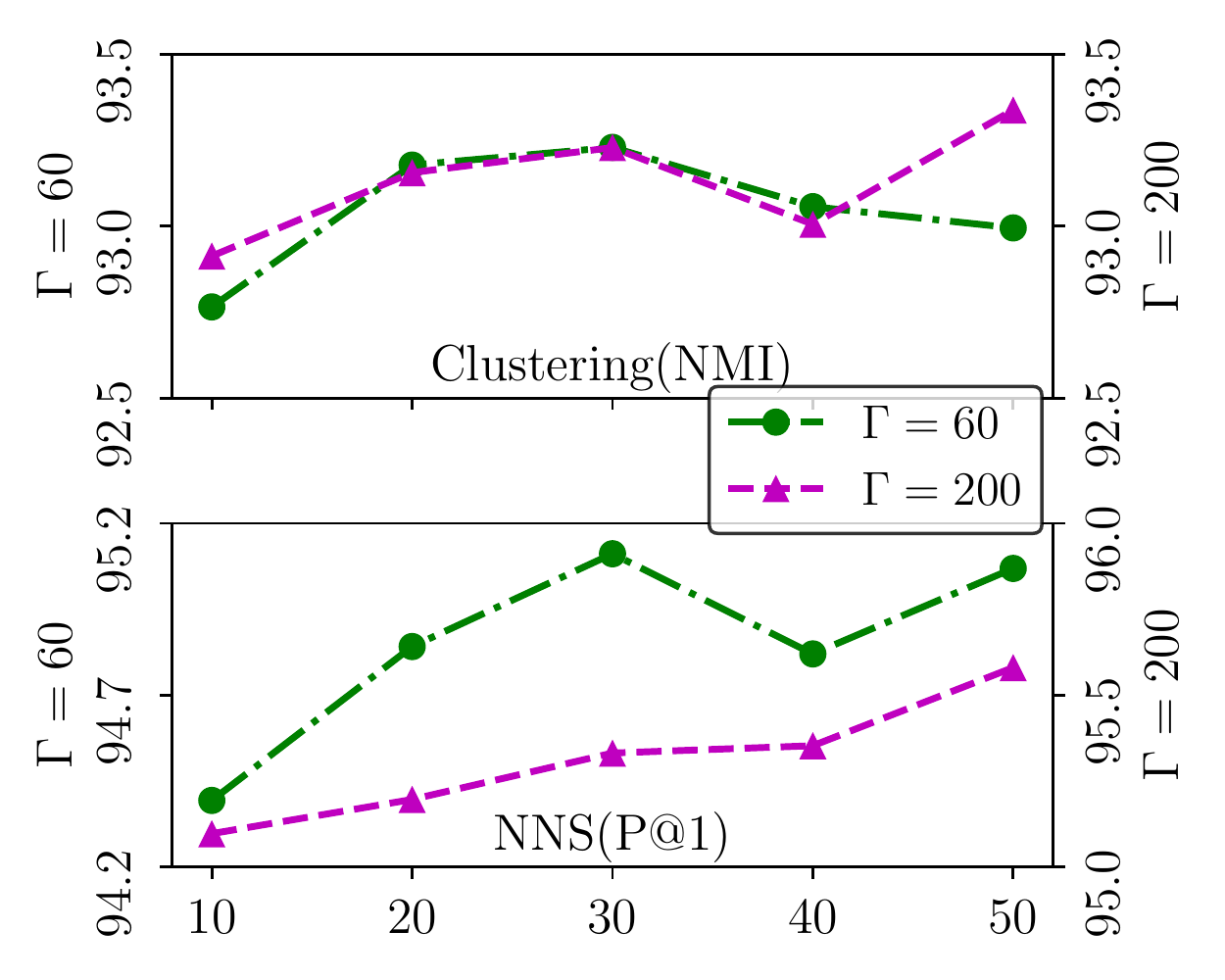}
  \caption{DE-SynPUF.} \label{fig:varied_attn_synpuf}
\end{subfigure}
\caption{NMI and P@1 (\%) on NUH2012 and DE-SynPUF by varying
temporal context scopes from 10 to 50.}
\label{fig:varied_attn}
\end{figure}

In this work, we propose a temporal scope, and learn attentions for time units within this scope. We investigate the effects of the temporal scope in this section. Figure~\ref{fig:varied_attn} shows the results of MCE on two datasets with the temporal context scope from 10 to 50.
In each sub-figure, we report the performance trained with two different context threshold, i.e., $\Gamma=60$ and $\Gamma=200$ respectively. We note that when $\Gamma$ is 200, the number of contexts are mainly limited by the temporal scopes. Due to the different ranges of the results, We use two vertical axes for different values of $\Gamma$. 

As shown in Figure \ref{fig:varied_attn_nuh}, on NUH2012, both NMI and P@1 get
the highest values when the temporal scope is 30 (weeks). Considering that
NUH2012 only contains EMR data spanning a year, a 30-week temporal scope is wide
enough to cover most EMRs. Nevertheless, the performance decreases when the
temporal scope is larger than 30, which could be for the reason of data
sparsity. On the contrary, on DE-SynPUF, we observe that MCE achieves higher NMI
and P@1 as the temporal context scope and $\Gamma$ get larger, as shown in
Figure \ref{fig:varied_attn_synpuf}, which demonstrates its potentiality on large datasets.

\subsection{Qualitative Examples and Visualization}

\begin{figure}
  \centering
\includegraphics[width=0.8\linewidth]{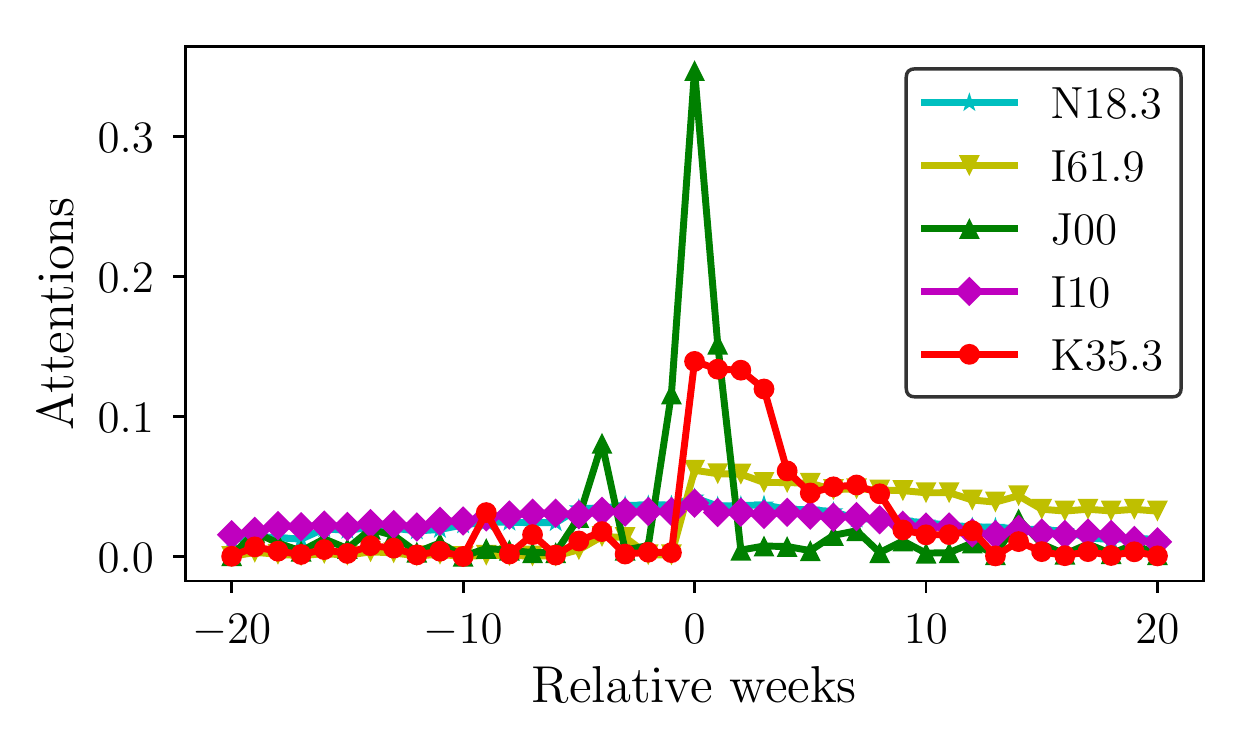}
\caption{``Soft'' time scopes of sample medical concepts. N18.3:
chronic kidney disease; I61.9: nontraumatic intracerebral hemorrhage; J00:
common cold; I10: essential hypertension; K35.3: acute appendicitis.}
\label{fig:soft_attn}
\end{figure}

Figure \ref{fig:soft_attn} shows the time-aware
attentions of sample medical concepts trained on NUH2012. The temporal scope spans 20 weeks before and
after the appearance of the medical concepts. With the help of the doctor, we
categorize the attention patterns into three types:

\begin{itemize}

  \item \textit{Stable influence.} Medical concepts such as \textit{chronic
  kidney disease} and \textit{essential hypertension}, which are associated with
  chronic diseases and complicated conditions, should have stable influence on
  every time period.

  \item \textit{Peak influence.} Some medical concepts are associated with acute
  diseases that could be recovered in a short time, such as \textit{common cold}
  and \textit{appendicitis}. They only have a peak influence on a short time
  span. 

  \item \textit{Sequela influence.} Some diseases have rare portent, but they
  are severe, such as \textit{nontraumatic intracerebral hemorrhage}. Such
  patients need a long time to recover. These medical concepts have notable
  influence on subsequently long-term periods.

\end{itemize}

